# “Be My Cheese?”: Cultural Nuance Benchmarking for Machine Translation in Multilingual LLMs

**Madison Van Doren[1,2], Casey Ford[1], Jennifer Barajas[1], Riley VanMeter[1,3], and Cory Holland[1]**

[1]Appen, [2]The University of Chicago, [3]The University of California, Berkeley

**Correspondence:** mvandoren@uchicago.edu, cford@appen.com

## Abstract

We present a large-scale human evaluation benchmark for assessing cultural localisation in machine translation produced by state-of-the-art multilingual large language models (LLMs). Existing MT benchmarks emphasise token-level and grammatical accuracy, but often overlook the pragmatic and culturally grounded competencies required for real-world localisation. Building on a pilot study of 87 translations across 20 languages, we evaluate 7 multilingual LLMs across 15 target languages with 5 native-speaker raters per language. Each rater scored both full-text translations and segment-level instances of culturally nuanced language (idioms, puns, holidays, and culturally embedded concepts) on an ordinal 0–3 quality scale; segment ratings additionally included an NA option for untranslated segments.

Across full-text evaluations, mean overall quality is modest (1.68/3): GPT-5 (2.10/3), Claude Sonnet 4 (1.97/3), and Mistral Medium 3.1 (1.84/3) form the strongest tier with fewer catastrophic failures. Segment-level results show sharp category effects: holidays (2.20/3) and cultural concepts (2.19/3) translate notably better than idioms (1.65/3) and puns (1.45/3), and idioms are most likely to be left untranslated. Inter-rater reliability was assessed using Krippendorff's α and Gwet's AC2, indicating moderate agreement overall (Krippendorff's $\alpha = 0.45$) with the lowest agreement for puns. These findings demonstrate a persistent gap between grammatical adequacy and cultural resonance. To our knowledge, this is the first multilingual, human-annotated benchmark focused explicitly on cultural nuance in translation and localisation. The results highlight the need for culturally informed training data, improved cross-lingual pragmatics, and evaluation frameworks that support systematic benchmarking of culturally grounded translation.

## 1. Introduction

Large language models (LLMs) have rapidly expanded access to machine translation, enabling rapid translation across hundreds of languages without requiring linguistic expertise. Cultural nuances, such as figurative expressions and idioms, are foundational to effective human communication and shape how meaning is received and interpreted. A translation that is grammatically correct may nevertheless sound unnatural, inappropriate, or misleading if it fails to account for cultural context. For example, in the present study, the pun "will you brie mine?" was frequently translated as, "be my cheese", a translation that is grammatically valid yet strips away both the romantic allusion and the wordplay that made the original effective. Despite this, machine translation (MT) research and benchmarks continue to prioritise lexical and grammatical accuracy at the token- and sentence-level. These metrics capture formal correctness, but fail to evaluate the pragmatic, cultural, and stylistic competencies required for real-world localisation tasks such as marketing communication, customer service, and culturally-specific messaging.

This study introduces a benchmark designed explicitly for evaluating how well multilingual LLMs preserve cultural resonance in machine translation tasks. Building on a pilot evaluation of 87 translations across 20 languages (Van Doren and Holland, 2025), we scale to a substantially larger dataset comprising 7 state-of-the-art multilingual LLMs, 15 target languages, and five native-speaker raters per language. Each rater evaluated both (1) a complete translated commercial email and (2) pre-defined segment-level instances of culturally nuanced language, including idioms, puns, holiday references, and culturally embedded concepts. This design allows us to contrast holistic translation quality with categorical failure modes on a phrasal level.

This benchmark enables systematic evaluation of culturally grounded translation, providing both holistic and fine-grained signals that are not captured by existing MT metrics. Our study addresses three core research questions:

- How well do contemporary multilingual LLMs translate culturally nuanced language across typologically diverse languages?
- To what extent do model family, linguistic characteristics, and orthographic systems impact cultural resonance in MT?
- Which categories of culturally marked content, such as idioms and puns, pose the greatest challenges to current LLMs?

Our findings reveal a substantial gap between grammatical accuracy and cultural localisation. While many translations achieve surface-level adequacy, even the strongest multilingual LLMs fail to consistently preserve culturally grounded meaning, particularly for figurative and non-literal language. These results underscore the limitations of existing machine translation in SOTA models and motivate a reevaluation of MT benchmarks and training practices that prioritise cultural–pragmatic competence as a core dimension of multilingual LLM performance.

### 1.1 Contributions

This work presents three primary contributions:

1. A new benchmark for culturally sensitive machine translation.

We introduce a multilingual, human-annotated benchmark for evaluating cultural localisation in machine translation, spanning 7 state-of-the-art multilingual LLMs, 15 languages, and five native-speaker raters per language. The benchmark combines full-text evaluation with segment-level annotation of culturally marked language, enabling both holistic assessment and fine-grained analysis of failure modes. All necessary materials to utilise this benchmark are publicly available at https://github.com/puzzlegoblin/fromage-benchmark/releases/tag/v1.0.

2. A large-scale empirical analysis of cultural failure modes in MT.

Through segment-level evaluation of idioms, puns, holidays, and culturally embedded concepts, we show that cultural localisation quality diverges sharply from grammatical accuracy, with figurative language remaining a persistent failure mode across models and languages.

3. Evidence of systematic model- and language-level variation in cultural MT performance.

We identify consistent performance differences across models, languages, and orthographic systems, including higher stability among GPT-5, Claude Sonnet 4, and Mistral Medium 3.1, and elevated failure rates for culturally marked segments in other systems, motivating targeted data and evaluation strategies for improving cultural competence in multilingual LLMs.

## 2. Related Work

Recent advances in LLMs have driven substantial improvements in multilingual machine translation. Mujadia et al. (2024) provide a comprehensive assessment of LLM translation performance between English and 22 Indian languages, revealing persistent disparities across high- and low-resource settings and demonstrating the benefits of in-context learning for underrepresented dialects. Similarly, Hu et al. (2024) introduce GenTranslate, showing that generative LLM-based approaches improve multilingual speech and text translation on standard benchmarks, particularly for low-resource languages. Together, these studies illustrate rapid progress in multilingual MT while highlighting uneven gains across languages.

Despite these advances, most prior evaluations focus on lexical and grammatical accuracy, relying on automatic metrics or sentence-level adequacy judgments. Such evaluations are poorly suited to capture pragmatic and cultural dimensions of translation quality, including idiomatic meaning, figurative language, and audience-appropriate tone. As a result, translations that are formally correct may nevertheless be culturally inappropriate or misleading in real-world localisation contexts. This limitation is well documented in the MT evaluation literature. BLEU has long been shown to correlate weakly with meaning adequacy and human judgments beyond surface correspondence (Callison-Burch et al., 2006; Mathur et al., 2020), and more recent neural metrics such as COMET and BLEURT similarly struggle with discourse-level, pragmatic, and culturally grounded errors (Freitag et al., 2021; Kocmi et al., 2022). Recent work has therefore begun to frame cultural transfer and adaptation as a core challenge for language technologies, arguing that culture-aware evaluation is necessary to capture meaning beyond surface correspondence (Singh et al., 2024). Relatedly, Stap et al. (2024) show that fine-tuning LLMs on parallel source-translation data improves COMET scores while simultaneously degrading formality steering, few-shot domain adaptation, and document-level contextualization, underscoring that gains on standard metrics can mask losses in the pragmatic competencies relevant to localisation.

Beyond translation accuracy, a growing body of work has examined cultural alignment in LLM outputs more broadly. AlKhamissi et al. (2024) investigate cultural alignment across languages and regions, showing that LLMs better reflect culturally grounded knowledge when prompted in a region's dominant language, while also identifying persistent representation gaps. Li et al. (2024) propose CultureLLM, incorporating culturally diverse multilingual data to improve cultural appropriateness in generation tasks. While these approaches demonstrate measurable gains, they largely focus on open-ended generation rather than translation and do not systematically evaluate how well models preserve culturally meaningful content when transferring meaning across languages.

The present work builds most directly on a pilot study by Van Doren and Holland (2025), which evaluated 87 translations across 20 languages and found that figurative language posed a consistent challenge even for high-performing models. While the pilot demonstrated the limitations of existing MT benchmarks for real-world localisation, it was constrained in scale and statistical power. The current study substantially extends this work by evaluating seven state-of-the-art multilingual LLMs across fifteen languages with multiple native-speaker raters per language, introducing segment-level evaluation of culturally nuanced language, and applying statistical modeling to disentangle the effects of model, language, and content type.

By situating cultural nuance as a core dimension of translation quality rather than a peripheral concern, this work complements existing MT and cultural alignment research and addresses a critical gap in current evaluation paradigms for multilingual LLMs.

## 3. Methodology

We evaluate multilingual LLMs on their ability to translate and culturally localise English commercial emails into 15 target languages. Unlike traditional MT benchmarks that emphasise lexical and grammatical accuracy, this task requires models to handle culturally marked language, including idioms, puns, holiday references, figurative expressions, and culturally embedded concepts.

Each model received the same English source text and a fixed prompt to, *"Translate the following email for use in [language] in [country/region]."* All translations were generated in fresh chat sessions to minimise contamination across runs.

### 3.1 Languages and Participants

This study included five native speakers per language (N = 75 total) across 15 locales: Afrikaans (ZA), Arabic (EG), Brazilian Portuguese (BR), Cantonese (HK), Czech (CZ), Dutch (NL), Hebrew (IL), Hindi (IN), Japanese (JP), Korean (KR), Mandarin (TW), Russian (KZ), Spanish (MX), Swahili (KE), and Urdu (PK).

Participants reside in the region they evaluated and are fluent speakers of both English and their native language. Each rater evaluated translations only for their native language (n= 5). Raters were compensated at a rate above local minimum wage according to standard internal practices. Participant demographic information is provided in Appendix B1.

### 3.2 Models Evaluated

We evaluate seven publicly available multilingual LLMs including a range of leading developers as well as open- and closed-weight systems.

| Developer | Model | Weight Type |
|---|---|---|
| Anthropic | Claude Sonnet 4 | Closed-weight |
| Mistral | Medium 3.1 | Closed-weight |
| DeepSeek | V3.1 | Closed-weight |
| OpenAI | GPT-5 | Closed-weight |
| OpenAI | gpt-oss 120B | Open-weight |
| Meta | Llama 4 | Open-weight |
| Cohere | Aya Expanse 8B | Open-weight |

Table 1: List of models evaluated.

When models produced meta-comments or explanations, English explanatory text was removed prior to evaluation. Non-English explanatory text was retained only when it was inseparable from the translated output and raters were instructed to ignore any explanatory output. Exact contributor instructions are available in Appendix B2.

### 3.3 Input Materials

Source texts consisted of five emails adapted from authentic commercial campaigns distributed between 2012-2020, prior to the public launch of ChatGPT to ensure the text was human-generated. These emails were selected to systematically elicit culturally marked language in realistic communicative settings. The marketing domain was chosen because it naturally combines persuasive intent, audience targeting, and frequent use of idiomatic and culturally specific expressions, making it a high-density setting for evaluating localisation quality.

From each email, we selected five segments of culturally nuanced language. Across the dataset, this resulted in four puns, four idioms, four holiday references, and thirteen cultural concepts per language. Cultural concepts were defined as single words or short phrases that are either specific to North American English or unlikely to have direct equivalents across cultures (e.g., koozies, sweetheart, zero-waste). Full source texts and segment selections are provided in Appendix A and at https://github.com/puzzlegoblin/fromage-benchmark/releases/tag/v1.0.

### 3.4 Evaluation Procedure

Each rater assessed one translation per model, evaluating both the full translated text and segments. A within-subjects design was employed wherein all participants rated all of the same segments for all models, within the context of the entire translation. Full participant guidelines are presented in Appendix B2 and at https://github.com/puzzlegoblin/fromage-benchmark/releases/tag/v1.0.

**(a) Full text evaluation**
Participants scored the translation on a 4-point scale for the following criteria:
1. Content fidelity
2. Style fidelity
3. Audience appropriateness
4. Overall translation quality

These items measure whether the translation is correct, natural, locally resonant, and aligned with the original intent. Participants were also given free response text boxes to provide additional qualitative feedback. A summary of the qualitative feedback by language is available in Appendix D.

**(b) Segment-level evaluation**
Raters also evaluated predefined culturally nuanced segments from the emails, each labeled as one of:

- idioms
- puns
- holidays
- cultural concepts

Segments were rated on the same 0–3 scale with an additional NA option to indicate when the segment was not translated and instead retained the original English. This enables fine-grained analysis of where models succeed or fail in cultural MT beyond full-text impressions. This methodology produced 13,125 segment-specific annotations.

### 3.5 Annotation Protocol

Participants received detailed written instructions based on an evaluation framework (available in Appendix B2), including:

- definitions of cultural nuance
- examples of literal vs. localised translation strategies
- guidance on how to rate ambiguous cases
- clarifications for rating idioms and humour

Ratings were collected using our proprietary data annotation software. Each submission was checked for completeness and annotation consistency.

### 3.6 Statistical Analysis

We analysed segment-level translation ratings using a cumulative link mixed model (CLMM) with a logit link, appropriate for ordinal outcomes. Models were fitted in R using the ordinal package (Christensen, 2022). Fixed effects included model, language, and segment category, as well as their interaction. Random intercepts were included for annotator and segment to account for repeated ratings and item-level variability.

Orthography was initially included as a fixed effect but was removed from the final specification due to rank deficiency and near-complete collinearity with language–category combinations. Its inclusion resulted in unstable parameter estimates without improving model fit. The final model converged successfully (logLik = −14,411.63; AIC = 28,965.26; n = 13,125). Random effects estimates indicate greater variance at the segment level (SD= 1.76) than at the annotator level (SD = 0.70), suggesting that segment-specific difficulty contributes more to rating variability than individual rater severity.

Inter-rater reliability (IRR) was assessed separately for full-text (overall) ratings and segment-level ratings using Krippendorff's α (ordinal) and Gwet's AC2 with quadratic weights. IRR was computed overall and stratified by model, language, and segment category. Full-text IRR assesses consistency in holistic translation judgments, while segment-level IRR evaluates agreement on fine-grained, culturally marked language. Ratings corresponding to "segment not translated" were excluded from IRR analyses.

Overall segment-level agreement was moderate (Krippendorff's α = 0.45; Gwet's AC2= 0.41), with lower values for puns (α = 0.31) and holidays (α = 0.38) than for cultural concepts (α = 0.44) and idioms (α = 0.40). Full IRR tables stratified by model, language, and segment category are reported in Appendix C.5.

## 4. Results

We report results from both holistic full-text evaluation and segment-level evaluation of culturally nuanced language. All scores are reported on a 0–3 ordinal scale, where higher values indicate better translation quality.

### 4.1 Full-Text Translation Quality by Model

Full-text translation quality remains modest overall (mean = 1.68/3). Descriptive averages place GPT-5 (2.10/3), Claude Sonnet 4 (1.97/3), and Mistral Medium 3.1 (1.84/3) at the top of the distribution, with Aya Expanse 8B substantially lower (1.09/3). Table 2 and Figure 1 summarise average full-text scores by model.

| Model | overall quality rating | audience | style | content |
|---|---|---|---|---|
| Claude Sonnet 4 | 1.97 | 2.25 | 2.08 | 2.10 |
| Cohere Aya Expanse 8B | 1.09 | 1.55 | 1.41 | 1.21 |
| DeepSeek V3.1 | 1.72 | 2.05 | 1.98 | 1.77 |
| GPT-5 | 2.10 | 2.38 | 2.23 | 2.23 |
| gpt-oss 120B | 1.60 | 1.94 | 1.83 | 1.72 |
| Llama 4 | 1.47 | 1.81 | 1.72 | 1.59 |
| Mistral Medium 3.1 | 1.84 | 2.19 | 2.04 | 1.92 |
| Total | 1.68 | 2.02 | 1.90 | 1.79 |

Table 2: Average rating on a 0–3 (4-point) ordinal scale by model across languages of overall translation quality, appropriateness to intended audience, faithfulness to style of the original, and faithfulness to content of the original.

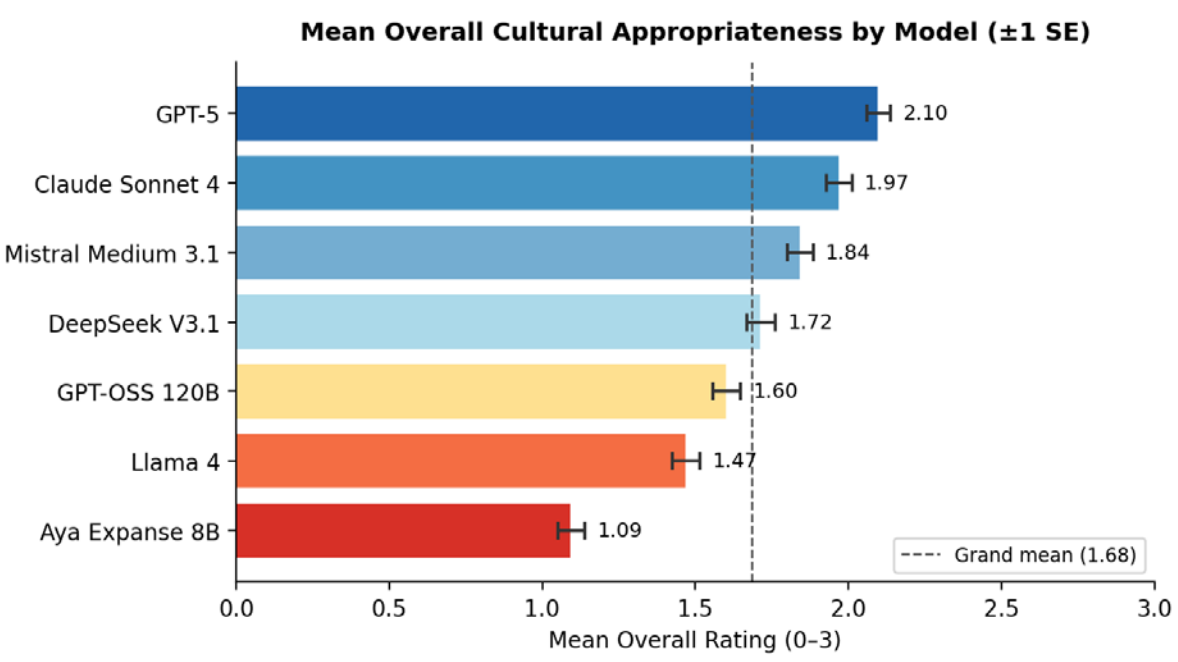

Figure 1: Mean overall cultural appropriateness ratings by model (±1 SE), collapsed across all 15 target languages.

CLMM results confirm a significant main effect of model on translation quality (Table C1). Relative to GPT-5, Aya Expanse 8B exhibits markedly worse performance ($\beta = 1.90$, $p < .001$). Llama 4, gpt-oss 120B, and DeepSeek V3.1 also perform significantly worse than GPT-5, while differences between GPT-5, Claude Sonnet 4, and Mistral Medium 3.1 are not statistically significant.

Estimated marginal means and Tukey-adjusted comparisons (Tables C2–C3) place GPT-5, Claude Sonnet 4, and Mistral Medium 3.1 as the highest performers, followed by a middle tier of DeepSeek V3.1 and gpt-oss 120B. Aya Expanse 8B is a clear outlier, performing significantly worse than all other models.

Inter-rater reliability for full-text ratings indicates moderate agreement across models and languages, supporting the stability of the observed model-level effects (Table C9). We report IRR to contextualise the subjectivity of cultural judgments while model and category effects are interpreted primarily through the CLMM estimates and post-hoc comparisons.

### 4.2 Segment-Level Performance by Category

Segment category exhibits the strongest and most consistent effect on translation quality. CLMM estimates show large and highly significant differences across categories (Tables C6–C7). Holidays and culturally embedded concepts receive substantially higher ratings than idioms and puns ($p < .001$ for all figurative vs. non-figurative contrasts), while the difference between idioms and puns is not statistically significant.

Descriptively, holidays (2.20/3) and cultural concepts (2.19/3) achieve the highest average quality among translated segments, whereas idioms (1.65/3) and puns (1.45/3) perform substantially worse. These results indicate that figurative and non-literal language remains a persistent challenge even when models attempt a translation.

Translation coverage also varies markedly by category. Idioms are most frequently left untranslated (rated NA), followed by puns, while holidays and cultural concepts are more consistently rendered. These omission patterns are reported descriptively and are not included in the CLMM, which models translation quality conditional on a translation being produced.

Segment-level IRR exhibit greater variability in inter-rater agreement, with lower agreement for puns and holidays than for idioms and cultural concepts, reflecting greater annotator uncertainty when evaluating figurative language (Table C8).

### 4.3 Model Effects on Segment Translation Quality

Controlling for language and segment category, model choice significantly affects segment-level translation quality (Table C1). While GPT-5 and Claude Sonnet 4 do not differ significantly, both show notable performance improvements when compared to gpt-oss 120B, Llama 4, and Aya Expanse 8B. Mistral Medium 3.1 performs significantly better than Aya Expanse 8B and Llama 4, but does not differ significantly from DeepSeek V3.1 or GPT-5.

Aya Expanse 8B is a clear outlier, exhibiting both significantly lower quality scores and substantially higher omission rates for idioms and puns. Other models omit fewer segments overall but frequently produce low-quality translations (ratings 0–1) for figurative language.

IRR stratified by model (Table C8) indicates moderate agreement for GPT-5 and Claude Sonnet 4, with greater variability for lower-performing models, suggesting that inconsistent output quality contributes to annotator disagreement.

### 4.4 Language-Level Effects

Language effects are present but more constrained than category or model effects: (Figure 2). CLMM estimates indicate that Mandarin (Taiwan) receives significantly higher segment-level ratings than several other languages, including Spanish, Swahili, and Urdu (Tables C4–C5). Brazilian Portuguese trends higher but does not consistently differ from other languages after correction for multiple comparisons.

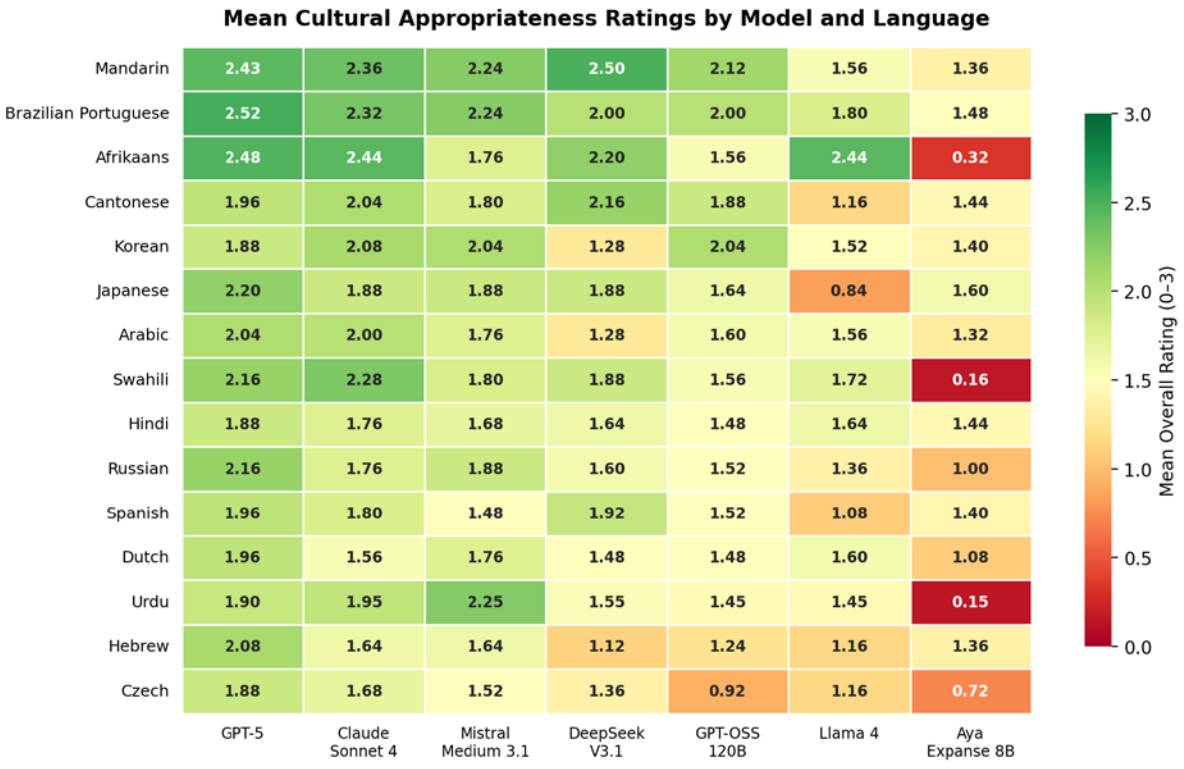

Figure 2: Mean overall cultural appropriateness ratings (0–3) by model and target language. Ratings reflect human judgments from five native speakers per language. Cells are ordered by descending mean rating across models (top) and languages (left). Notable variation is observed for Cohere Aya Expanse 8B, which performs substantially below other models for low-resource languages including Swahili (0.16), Urdu (0.15), and Afrikaans (0.32).

Importantly, language effects interact with segment category. Languages that perform well overall tend to maintain higher scores across categories, while lower-performing languages exhibit disproportionate degradation on idioms and puns. This pattern persists even when restricting analysis to translated segments, indicating that low scores are not driven solely by omission.

Language-level IRR ranged from $\alpha \approx 0.27$ (Hindi, Mandarin) to $\alpha \approx 0.57$ (Dutch), with lower values concentrated in languages where raters flagged greater cultural interpretive variability in qualitative feedback. This suggests that cultural interpretation differences may amplify annotator variability in these contexts.

## 5. Discussion

The CLMM analysis confirms that cultural localisation failures in multilingual LLMs are systematic rather than anecdotal. Segment category emerges as the strongest predictor of translation quality, exceeding the influence of both model family and language. Figurative language, especially idioms and puns, remains a robust failure mode even after controlling for rater effects and segment-level difficulty.

Crucially, the statistical results support a distinction between translation coverage and translation quality. Idioms are significantly more likely to be omitted entirely, and when translated, they receive substantially lower ratings than holidays or culturally embedded concepts. Aya Expanse 8B exhibits both the highest omission rates and the lowest quality scores for idiomatic translation, indicating that failure is not merely a consequence of conservative behavior. Even when models attempt figurative translation, pragmatic and culturally appropriate rendering frequently fails.

Model-level effects reveal a stable top tier (GPT-5, Claude Sonnet 4, and Mistral Medium 3.1) but no system consistently achieves high performance across all categories. The absence of statistically significant differences among these models suggests that scaling and architectural refinement alone do not resolve cultural–pragmatic limitations. In contrast, Aya Expanse 8B's consistently poor performance across analyses points to systemic fragility rather than isolated weaknesses.

Language-level effects are present but secondary, and orthography does not independently predict translation quality once language and segment category are taken into account. This finding contrasts with observations from the pilot study and challenges assumptions that script or typological complexity are primary drivers of cultural MT difficulty. Instead, the results point toward the availability and quality of culturally situated training data as a more plausible explanation for observed disparities.

## 6. Future work

Future work will extend this benchmark in several directions. We plan to expand the benchmark beyond text-only translation by developing an audio-based version of the task. This expansion will also include increasingly diverse contexts, such as machine translation of natural dialogue and non-commercial input texts.

Many culturally marked expressions (such as humour, idioms, and tone) are realised differently in spoken language, and evaluating speech-based localisation will allow analysis of dialect, prosody, emphasis, and pragmatic delivery not captured in text. We also intend to extend coverage to additional domains and languages to assess the generality of the cultural failure modes identified here. Further work will also analyse the effectiveness and appropriateness of omitting translation of a segment (retaining the English).

## 7. Conclusion

We presented a large-scale, human-annotated benchmark designed to evaluate cultural localisation in machine translation by multilingual LLMs. Across seven state-of-the-art models and fifteen languages, results reveal a persistent gap between grammatical adequacy and cultural resonance. While many translations appear superficially plausible, segment-level evaluation exposes systematic failures, particularly idioms and puns, that remain largely invisible to standard MT metrics.

By explicitly distinguishing between translation coverage and translation quality, this work provides a more nuanced account of cultural MT performance and highlights limitations shared even by the strongest current models. These findings underscore the need for culturally informed training data and evaluation paradigms that move beyond form-based correctness toward real-world communicative competence.

## 8. Data and Code Availability

We release the dataset and evaluation framework as a public benchmark, enabling reproducible research on cultural localisation in machine trans-lation and multilingual LLM evaluation. The release includes full-text translations, segment-level annotations, and detailed evaluation guidelines to support consistent comparison across future models. Rather than replacing automatic metrics, this benchmark complements them by targeting pragmatic and cultural dimensions that current form-based evaluations systematically overlook.

Beyond identifying failure modes, this benchmark provides a framework for systematically evaluating cultural competence in multilingual LLMs. First, it provides a standardised human evaluation protocol for comparing multilingual LLMs on culturally grounded translation tasks, complementing existing form-based metrics. Second, the segment-level annotations enable targeted error analysis and category-specific evaluation (e.g., idioms vs. puns), which can be used to diagnose model weaknesses and guide data curation or fine-tuning. Third, the released annotations can serve as supervision for developing learned evaluation metrics that better capture cultural and pragmatic adequacy, addressing known limitations of current automatic metrics. By enabling controlled comparison across models, languages, and categories of culturally marked language, it supports both diagnostic evaluation and the development of improved metrics and training strategies. We hope this work contributes to a shift toward evaluation paradigms that better reflect real-world communicative effectiveness.

## Limitations

Limitations of this study include the benchmark's focus on English-to-many translation within a commercial email domain, which may limit generalisability to other genres. Segment selection intentionally emphasises culturally marked language and is therefore not representative of typical sentence distributions in MT corpora. Furthermore, analysis of segment-level MT was performed in the context of larger MT corpora and may not generalise to MT performance when translating the same segments as isolated text. Future work could rectify this limitation by evaluating and contrasting segment-level MT in context of larger text with segment MT as isolated input.

Although five native raters per language reduce individual bias, judgments of cultural appropriateness remain inherently subjective and may vary across demographics, regions, and personal experience. Additionally, this study did not control for differences in participant age, education, gender, and socioeconomic background – all factors known to influence human bias (Jenks, 2025; Zahraei and Emami, 2025). In addition, models were evaluated through publicly accessible interfaces, which may introduce uncontrolled variation due to system prompts, safety filters, or model updates. Furthermore, model outputs were collected over the span of two days, introducing additional potential for uncontrolled variation when compared to simultaneous output generation and collection. Finally, this work focuses exclusively on text-based translation and does not address multimodal or spoken localisation, which we leave to future research.

## Ethics Statement

This study involved human participants as cultural appropriateness raters for machine-translated text. All participants were recruited through CrowdGen, Appen's contributor platform, and provided informed consent prior to participation. The study protocol did not involve identifiable personal data and posed no risk to participants.

Participants were compensated at a rate commensurate with or above the applicable local minimum wage in their region. No deception was involved; participants were informed of the nature of the task before beginning.

The source email texts used as translation stimuli were drawn from marketing emails sent to the first author's personal email account and contain no personally identifiable information. Demographic data collected from annotators (language background, country of residence, age group, and education level) were used solely for reporting participant characteristics and subgroup analysis, and are reported only in aggregate.

The models evaluated in this study are commercially available systems accessed through public APIs and interfaces. No fine-tuning, model extraction, or adversarial probing was performed. The benchmark and annotations are released to support further research on culturally sensitive machine translation evaluation; we encourage downstream users to consider potential harms of deploying MT systems in culturally high-stakes contexts without human review.

## References

Badr AlKhamissi, Muhammad ElNokrashy, Mai Alkhamissi, and Mona Diab. 2024. Investigating Cultural Alignment of Large Language Models. In *Proceedings of the 62nd Annual Meeting of the Association for Computational Linguistics (Volume 1: Long Papers)*, pages 12404–12422, Bangkok, Thailand. Association for Computational Linguistics.

Chris Callison-Burch, Miles Osborne, and Philipp Koehn. 2006. Re-evaluating the Role of Bleu in Machine Translation Research. In *11th Conference of the European Chapter of the Association for Computational Linguistics*, pages 249–256, Trento, Italy. Association for Computational Linguistics.

Rune Haubo Bojesen Christensen. 2022. ordinal: Regression models for ordinal data**.** R package.

Markus Freitag, George Foster, David Grangier, Viresh Ratnakar, Qijun Tan, and Wolfgang Macherey. 2021. Experts, Errors, and Context: A Large-Scale Study of Human Evaluation for Machine Translation. *Transactions of the Association for Computational Linguistics*, 9:1460–1474.

Yuchen Hu, Chen Chen, Chao-Han Huck Yang, Ruizhe Li, Dong Zhang, Zhehuai Chen, and Eng Siong Chng. 2024. GenTranslate: Large language models are generative multilingual speech and machine translators. In *Proceedings of the 62nd Annual Meeting of the Association for Computational Linguistics*. Association for Computational Linguistics.

Christopher Jenks. 2025. Communicating the cultural other: Trust and bias in generative AI and large language models**.** *Applied Linguistics Review*, 16(2):787–795.

Klaus Krippendorff. 2019. Content analysis: An introduction to its methodology. Sage.

Cheng Li, Mengzhuo Chen, Jindong Wang, and Sunayana Sitaram. 2024. CultureLLM: Incorporating cultural differences into large language models. In *Proceedings of the 38th Conference on Neural Information Processing Systems*.

Nitika Mathur, Timothy Baldwin, and Trevor Cohn. 2020. Tangled up in BLEU: Reevaluating the Evaluation of Automatic Machine Translation Evaluation Metrics. In *Proceedings of the 58th Annual Meeting of the Association for Computational Linguistics*, pages 4984–4997, Online. Association for Computational Linguistics.

Vandan Mujadia, Ashok Urlana, Yash Bhaskar, Penumalla Aditya Pavani, Kukkapalli Shravya, Parameswari Krishnamurthy, and Dipti Sharma. 2024. Assessing Translation Capabilities of Large Language Models involving English and Indian Languages. In *Proceedings of the 25th Annual Conference of the European Association for Machine Translation (Volume 1)*, pages 207–228, Sheffield, UK. European Association for Machine Translation (EAMT).

Pushpdeep Singh, Mayur Patidar, and Lovekesh Vig. 2024. Translating across cultures: LLMs for intralingual cultural adaptation. In *Proceedings of the 28th Conference on Computational Natural Language Learning*, pages 400–418, Miami, FL, USA. Association for Computational Linguistics.

David Stap, Eva Hasler, Bill Byrne, Christof Monz, and Ke Tran. 2024. The fine-tuning paradox: Boosting translation quality without sacrificing LLM abilities. In Findings of the Association for Computational Linguistics: ACL 2024, pages 6189–6206, Bangkok, Thailand. Association for Computational Linguistics.

Madison Van Doren & Cory Holland. 2025, "Be My Cheese?": Assessing Cultural Nuance in Multilingual LLM Translations, arXiv:2509.21577, Presented at the Linguistics Society of America Annual Meeting 2026, New Orleans.

Linting Xue, Aditya Barua, Noah Constant, Rami Al-Rfou, Sharan Narang, Mihir Kale, Adam Roberts, and Colin Raffel. 2022. ByT5: Towards a token-free future with byte-level models**.** *Transactions of the Association for Computational Linguistics*.

Pardis Sadat Zahraei and Ali Emami. 2025. Translate With Care: Addressing Gender Bias, Neutrality, and Reasoning in Large Language Model Translations. In *Findings of the Association for Computational Linguistics: ACL 2025*, pages 476–501, Vienna, Austria. Association for Computational Linguistics.

# Appendix

## Appendix A. Complete MT Input Texts

### A1

Company: Sheffield's – a gourmet market in NYC

Subject: Will you brie mine? 🧀❤️🧀

Valentine's Day is almost here, and we've got the sweetest gift ideas for pickup or delivery throughout NYC.

Cheese Tasting Gift Boxes

This cheese lover's dream is thoughtfully assembled by our expert cheesemongers. It all comes beautifully packaged in a keepsake tin, tied with a satin ribbon. Personalise it with a custom note on Sheffield's stationery.

Sweets for your Sweetheart

Artfully displayed with the perfect accompaniments of fresh & dried fruit, nuts, honey, fig jam, espresso brownies, dark chocolate-covered strawberries, candies, edible flowers and sliced baguette.

[order here]

We still have a limited number of handmade, chocolate-covered strawberries and floral arrangements available for pre-order! Give us a call today or stop by the shop before they're gone.

Wishing you a sweet Valentine's Day!

Sheffield's – Park Slope

Brooklyn, NY

### A2

Company: Terra – an eco-friendly deodorant brand

Subject: This scent will transform your life ✨

Hey [NAME]

Your New Year's resolution stinks. Give your life a scent-sational upgrade – pair our newest reusable case design with a fragrance that's sure to make memories. Durable, stylish, compact, and zero waste.

Swipe right this New Year's Eve

Use code: NYE2026

[shop deodorant]

Whether you're keeping yourself fresh for your partner, or looking to impress someone else, our new scents will leave a lasting impression.

MIX & MATCH OUR BEST-SELLING COMBOS

Lavender case x Tropical Paradise scent

Turquoise case x Orange Creamsicle scent

WHY TERRA?

Aluminum & paraben free. Zero-waste refills. 24-hour odor protection. All that in a case you'll be excited to reuse.

Terra Cosmetics

London N1C 4AB, United Kingdom

### A3

Company: Muggable – an American novelty mug company

Subject: This Collection Has Us Feline Good 🐱

CAT'S MEOW

Our newest collection is the cat's pajamas, wait no – it's the cat's Mugs, Tumblers, Koozies, and Coasters!

[Shop Meow]

Rep your favorite feline at the office, on the go, and on your next Zoom call. Wait. Who are we kidding? They're already in all your Zoom calls.

**A4**

Company: sonia summerhouse– an american luxury swimwear brand

Subject: late Summer, full throttle

Labor Day is here! PACK YOUR BEACH BAG!

you sprint barefoot across warm sand.

the sun hits high.

salt hangs in the air.

seagulls cut the wide blue sky.

laughter bursts, waves crash in time,

summer comes alive.
your new swimwear, green like sea glass.

fabric flowing, grab your crew,

chase the surf, leap, sprint, splash -

shore enough, this is your moment!

Sonia Summerhouse 20 w. 20th street unit 1004 new york, ny 10011

**A5**

company: Cinnamon – a neighborhood bakery & cafe

Subject: Happy Birthday! There's a sweet treat waiting for you!

Sugar, spice, and everything nice! Happy Birthday from all of us at Cinnamon!

Let us be the icing on the cake of your special day with a sweet treat. Stop by any Cinnamon location to redeem your credit on your next order OR save it for later by visiting the Rewards section in your app.

We can't wait to celebrate with you! Redeemable with the Cinnamon app only.

Excited about your birthday present?
Say Thanks on Facebook

### A6 Segments

| Segment | Segment category |
|---|---|
| birthday present | cultural concepts |
| cheesemongers | cultural concepts |
| full throttle | cultural concepts |
| grab your crew | cultural concepts |
| Happy Birthday | cultural concepts |
| keepsake tin | cultural concepts |
| Koozies | cultural concepts |
| summer comes alive | idioms |
| sweet treat | cultural concepts |
| Sweetheart | cultural concepts |
| Swipe right | cultural concepts |
| Tumblers | cultural concepts |
| Zero-waste | cultural concepts |
| Zoom call | cultural concepts |
| New Year’s Eve | holidays |
| NYE2026 | holidays |
| Labor Day | holidays |
| Valentine’s Day | holidays |
| cat’s pajamas | idioms |
| icing on the cake | idioms |
| Sugar, spice, and everything nice | idioms |
| Feline Good | puns |
| scent-sational | puns |
| shore enough | puns |
| Will you brie mine? | puns |

Table A6 Segmentation and categorisation of phrases and words selected for individual evaluation.

## Appendix B. Participants

### B1 Participant Demographics

| Language | Participant age | Participant Gender | Participant education level |
|---|---|---|---|
| Afrikaans | 31-45 | female | Secondary education completed (high school diploma or equivalent) |
| Afrikaans | 31-45 | female | Postgraduate diploma or certificate (non-degree) |
| Afrikaans | 31-45 | female | Postgraduate diploma or certificate (non-degree) |
| Afrikaans | 31-45 | female | Some college or university (no degree) |
| Afrikaans | 45+ | female | Some college or university (no degree) |
| Arabic | 45+ | female | Bachelor's degree (e.g., BA, BS) |
| Arabic | 18-30 | male | Bachelor's degree (e.g., BA, BS) |
| Arabic | 31-45 | male | Master's degree (e.g., MA, MS, MBA, MFA) |
| Arabic | 31-45 | female | Bachelor's degree (e.g., BA, BS) |
| Arabic | 18-30 | male | Bachelor's degree (e.g., BA, BS) |
| Brazilian Portuguese | 18-30 | male | Secondary education completed (high school diploma or equivalent) |
| Brazilian Portuguese | 31-45 | female | Some college or university (no degree) |
| Brazilian Portuguese | 45+ | male | Master's degree (e.g., MA, MS, MBA, MFA) |
| Brazilian Portuguese | 31-45 | male | Master's degree (e.g., MA, MS, MBA, MFA) |
| Brazilian Portuguese | 45+ | male | Postgraduate diploma or certificate (non-degree) |
| Cantonese | 31-45 | female | Bachelor's degree (e.g., BA, BS) |
| Cantonese | 18-30 | female | Bachelor's degree (e.g., BA, BS) |
| Cantonese |  | UNKNOWN | Bachelor's degree (e.g., BA, BS) |
| Cantonese | 18-30 | female | Bachelor's degree (e.g., BA, BS) |
| Cantonese | 45+ | female | Bachelor's degree (e.g., BA, BS) |
| Czech | 31-45 | female | Bachelor's degree (e.g., BA, BS) |
| Czech | 18-30 | female | Master's degree (e.g., MA, MS, MBA, MFA) |
| Czech | 18-30 | male | Some secondary education (high school) |
| Czech | 18-30 | male | Some college or university (no degree) |
| Czech | 31-45 | male | Master's degree (e.g., MA, MS, MBA, MFA) |
| Dutch | 31-45 | male | Bachelor's degree (e.g., BA, BS) |
| Dutch | 31-45 | male | Bachelor's degree (e.g., BA, BS) |
| Dutch | 45+ | female | Bachelor's degree (e.g., BA, BS) |
| Dutch | 31-45 | male | Bachelor's degree (e.g., BA, BS) |
| Dutch | 45+ | male | Bachelor's degree (e.g., BA, BS) |
| Hebrew | 45+ | male | Bachelor's degree (e.g., BA, BS) |
| Hebrew | 31-45 | male | Bachelor's degree (e.g., BA, BS) |
| Hebrew | 31-45 | female | Bachelor's degree (e.g., BA, BS) |
| Hebrew | 31-45 | male | Vocational/technical training or certification (e.g., trade school) |
| Hebrew | 45+ | male | Bachelor's degree (e.g., BA, BS) |
| Hindi | 31-45 | male | Bachelor's degree (e.g., BA, BS) |

| Hindi | 31-45 | male | Doctoral or professional degree (e.g., PhD, MD, JD, PsyD, EdD) |
|---|---|---|---|
| Hindi | 45+ | male | Master's degree (e.g., MA, MS, MBA, MFA) |
| Hindi | 18-30 | male | Postgraduate diploma or certificate (non-degree) |
| Hindi | 18-30 | male | Bachelor's degree (e.g., BA, BS) |
| Japanese | 45+ | female | Bachelor's degree (e.g., BA, BS) |
| Japanese | 31-45 | male | Master's degree (e.g., MA, MS, MBA, MFA) |
| Japanese | 31-45 | male | Bachelor's degree (e.g., BA, BS) |
| Japanese | 45+ | male | Bachelor's degree (e.g., BA, BS) |
| Japanese | 18-30 | male | Bachelor's degree (e.g., BA, BS) |
| Korean | 45+ | female | Bachelor's degree (e.g., BA, BS) |
| Korean | 31-45 | female | Master's degree (e.g., MA, MS, MBA, MFA) |
| Korean | | | Bachelor's degree (e.g., BA, BS) |
| Korean | 31-45 | female | Master's degree (e.g., MA, MS, MBA, MFA) |
| Korean | 45+ | female | Bachelor's degree (e.g., BA, BS) |
| Mandarin | 31-45 | female | Bachelor's degree (e.g., BA, BS) |
| Mandarin | 31-45 | female | Bachelor's degree (e.g., BA, BS) |
| Mandarin | 45+ | male | Master's degree (e.g., MA, MS, MBA, MFA) |
| Mandarin | 18-30 | male | Bachelor's degree (e.g., BA, BS) |
| Mandarin | 31-45 | male | Master's degree (e.g., MA, MS, MBA, MFA) |
| Russian | 31-45 | male | Master's degree (e.g., MA, MS, MBA, MFA) |
| Russian | 31-45 | female | Bachelor's degree (e.g., BA, BS) |
| Russian | 45+ | male | Secondary education completed (high school diploma or equivalent) |
| Russian | 45+ | female | Bachelor's degree (e.g., BA, BS) |
| Russian | 31-45 | male | Master's degree (e.g., MA, MS, MBA, MFA) |
| Spanish | 31-45 | female | Bachelor's degree (e.g., BA, BS) |
| Spanish | 31-45 | female | Bachelor's degree (e.g., BA, BS) |
| Spanish | 18-30 | female | Master's degree (e.g., MA, MS, MBA, MFA) |
| Spanish | 31-45 | FEMALE | Some college or university (no degree) |
| Spanish | 31-45 | male | Bachelor's degree (e.g., BA, BS) |
| Swahili | 18-30 | female | Bachelor's degree (e.g., BA, BS) |
| Swahili | 31-45 | female | Postgraduate diploma or certificate (non-degree) |
| Swahili | 18-30 | female | Bachelor's degree (e.g., BA, BS) |
| Swahili | 18-30 | male | Bachelor's degree (e.g., BA, BS) |
| Swahili | 18-30 | male | Bachelor's degree (e.g., BA, BS) |
| Urdu | 31-45 | male | Master's degree (e.g., MA, MS, MBA, MFA) |
| Urdu | 31-45 | female | Master's degree (e.g., MA, MS, MBA, MFA) |
| Urdu | 31-45 | female | Master's degree (e.g., MA, MS, MBA, MFA) |
| Urdu | 18-30 | male | Master's degree (e.g., MA, MS, MBA, MFA) |
| Urdu | 31-45 | male | Bachelor's degree (e.g., BA, BS) |

### B2 Participant Guidelines

**Overview**

This project is meant to evaluate the quality of translation and localization of various LLM models when asked to translate marketing emails from English to a given language and locale. Imagine that a person working at an advertising agency is asked to translate a marketing email they are working on from English to a language and country that they don't know anything about. They do what people do these days and go to the internet and ask their favorite LLM model to "Translate this email into {{language}} for use in {{country}}"

You represent their end user, as a person in the targeted country who speaks the language, we are asking for. We'd like you to evaluate the email from the perspective of a person getting that advertisement in your email. How well is it translated? How well does it target local traditions and norms? How true to the original content and tone is the translation"

We'll ask several questions, all using the same basic evaluation scale. Keep these ratings and descriptions in mind while you are evaluating.

- serious failures exist - use this in cases where you would be very disappointed, confused, offended, or in some other way have negative feelings towards the company or product because of the content of the translation
- imperfect but not terrible - there are errors or issues that are very noticeable, but that are not so bad as to give a negative impression of the company, the main ideas come through and it is clear what is being advertised.
- mostly good with small issues - the wording or translations are noticeably non-native, or are awkward or a little odd, but it is overall something that makes sense and could be used without embarrassment on the part of the company.
- very good or nearly perfect - this is for something that seems very close to natural, native, and culturally appropriate.

Steps

1. Read both emails.
2. Answer the holistic questions on the left hand side
3. Answer the segment specific questions on the right hand side
4. Give an overall evaluation, considering your ratings both for the holistic and the segment specific ratings
5. Leave free-text comments at any point if you notice something interesting or want to add context to your rating.

Notes

- The translated email may include notes from the model on the translation. Please disregard these and evaluate the translated email from the perspective of a potential customer receiving it in your inbox.

## Appendix C. Statistical Modelling

### C.1 Cumulative Link Mixed Model Specification

A cumulative link mixed model (CLMM) with a logit link function was fitted to the ordinal translation quality ratings using the ordinal package in R (Christensen, 2022). The model was specified to account for the ordered nature of the response variable while incorporating both fixed and random effects to capture systematic variation across experimental factors and repeated measurements.

Orthography was initially included as a fixed effect; however, preliminary diagnostics indicated that its inclusion resulted in a rank-deficient design matrix, with multiple coefficients automatically dropped during estimation. Inspection of the data revealed sparse cell counts and near-complete collinearity between orthography, language, and segment category. Under these conditions, orthography effects could not be uniquely identified and impeded stable estimation without improving model fit. Orthography was therefore excluded from the final model to preserve identifiability and convergence.

The final fixed-effects structure included model, language, segment category, and their interaction. Random intercepts were specified for annotator and segment to account for repeated ratings by the same individuals and shared difficulty across evaluated segments. Model parameters were estimated via maximum likelihood using the regularised Newton–Raphson algorithm implemented in ordinal.

| Predictor | Estimate | SE | CI | z_ratio | p_value | Significance |
|---|---|---|---|---|---|---|
| Very good / nearly perfect\|Mostly good | -0.01 | 0.36 | [-0.71, 0.69] | -0.03 | 0.975 | |
| Mostly good\|Imperfect | 1.28 | 0.36 | [0.58, 1.99] | 3.57 | < .001 | *** |
| Imperfect\|Serious failures | 2.50 | 0.36 | [1.80, 3.21] | 6.95 | < .001 | *** |
| Serious failures\|Segment not translated | 6.27 | 0.37 | [5.54, 6.99] | 16.92 | < .001 | *** |
| modelCohere Aya Expanse 8B | 1.90 | 0.15 | [1.60, 2.20] | 12.42 | < .001 | *** |
| modelDeepSeek V3.1 | 0.51 | 0.15 | [0.20, 0.81] | 3.28 | 0.001 | ** |
| modelGPT-5 | 0.02 | 0.16 | [-0.28, 0.33] | 0.15 | 0.878 | |
| modelgpt-oss 120b | 0.81 | 0.15 | [0.50, 1.11] | 5.25 | < .001 | *** |
| modelLlama 4 | 1.03 | 0.15 | [0.72, 1.33] | 6.68 | < .001 | *** |
| modelMistral Medium 3.1 | 0.38 | 0.15 | [0.08, 0.69] | 2.48 | 0.013 | * |
| languageArabic | 0.22 | 0.48 | [-0.72, 1.15] | 0.45 | 0.652 | |
| languageBrazilian Portuguese | -0.92 | 0.48 | [-1.87, 0.03] | -1.89 | 0.058 | |
| languageCantonese | -0.22 | 0.48 | [-1.15, 0.72] | -0.45 | 0.650 | |
| languageCzech | 0.12 | 0.48 | [-0.81, 1.05] | 0.25 | 0.800 | |
| languageDutch | 0.29 | 0.48 | [-0.64, 1.23] | 0.62 | 0.538 | |
| languageHebrew | 0.16 | 0.48 | [-0.77, 1.10] | 0.34 | 0.735 | |
| languageHindi | 0.60 | 0.47 | [-0.32, 1.53] | 1.28 | 0.202 | |
| languageJapanese | -0.29 | 0.48 | [-1.23, 0.64] | -0.61 | 0.540 | |
| languageKorean | 0.14 | 0.48 | [-0.81, 1.09] | 0.29 | 0.771 | |
| languageMandarin | -1.53 | 0.49 | [-2.50, -0.56] | -3.09 | 0.002 | ** |
| languageRussian | -0.53 | 0.49 | [-1.49, 0.44] | -1.07 | 0.284 | |
| languageSpanish | 0.17 | 0.47 | [-0.76, 1.09] | 0.35 | 0.726 | |
| languageSwahili | 0.18 | 0.48 | [-0.76, 1.11] | 0.37 | 0.711 | |
| languageUrdu | 0.33 | 0.49 | [-0.63, 1.28] | 0.67 | 0.501 | |
| segment_category.L | 1.66 | 0.08 | [1.49, 1.82] | 19.72 | < .001 | *** |
| segment_category.Q | 0.31 | 0.10 | [0.12, 0.49] | 3.22 | 0.001 | ** |
| segment_category.C | -0.84 | 0.11 | [-1.05, -0.63] | -7.79 | < .001 | *** |

Table C1 Fixed-effect estimates from the cumulative link mixed model predicting machine translation quality (0–3).

### C2 Model-Level Effects

| factor | emmean | SE | CI |
|---|---|---|---|
| Claude Sonnet 4 | -2.60 | 0.14 | [-2.87, -2.32] |
| Cohere Aya Expanse 8B | -0.69 | 0.14 | [-0.96, -0.43] |
| DeepSeek V3.1 | -2.09 | 0.14 | [-2.36, -1.82] |
| GPT-5 | -2.57 | 0.14 | [-2.85, -2.29] |
| gpt-oss 120B | -1.79 | 0.14 | [-2.06, -1.52] |
| Llama 4 | -1.57 | 0.14 | [-1.84, -1.30] |
| Mistral Medium 3.1 | -2.21 | 0.14 | [-2.49, -1.94] |

Table C2 Estimated Marginal Means by Model

| contrast | estimate | SE | CI | z_ratio | p_value | Significance |
|---|---|---|---|---|---|---|
| Claude Sonnet 4 - Cohere Aya Expanse 8B | -1.90 | 0.15 | [-2.36, -1.45] | -12.42 | < .001 | *** |
| Claude Sonnet 4 - DeepSeek V3.1 | -0.51 | 0.15 | [-0.96, -0.05] | -3.28 | 0.018 | * |
| Claude Sonnet 4 - (GPT-5) | -0.02 | 0.16 | [-0.48, 0.43] | -0.15 | 1.000 | |
| Claude Sonnet 4 - (gpt-oss 120b) | -0.81 | 0.15 | [-1.26, -0.35] | -5.25 | < .001 | *** |
| Claude Sonnet 4 - Llama 4 | -1.03 | 0.15 | [-1.48, -0.57] | -6.68 | < .001 | *** |
| Claude Sonnet 4 - Mistral Medium 3.1 | -0.38 | 0.15 | [-0.84, 0.07] | -2.48 | 0.165 | |
| Cohere Aya Expanse 8B - DeepSeek V3.1 | 1.40 | 0.15 | [0.95, 1.84] | 9.24 | < .001 | *** |
| Cohere Aya Expanse 8B - (GPT-5) | 1.88 | 0.15 | [1.43, 2.33] | 12.31 | < .001 | *** |
| Cohere Aya Expanse 8B - (gpt-oss 120b) | 1.10 | 0.15 | [0.66, 1.54] | 7.32 | < .001 | *** |
| Cohere Aya Expanse 8B - Llama 4 | 0.88 | 0.15 | [0.43, 1.32] | 5.84 | < .001 | *** |
| Cohere Aya Expanse 8B - Mistral Medium 3.1 | 1.52 | 0.15 | [1.07, 1.97] | 10.03 | < .001 | *** |
| DeepSeek V3.1 - (GPT-5) | 0.48 | 0.15 | [0.03, 0.94] | 3.13 | 0.029 | * |
| DeepSeek V3.1 - (gpt-oss 120B) | -0.30 | 0.15 | [-0.75, 0.15] | -1.97 | 0.432 | |
| DeepSeek V3.1 - Llama 4 | -0.52 | 0.15 | [-0.97, -0.07] | -3.42 | 0.011 | * |
| DeepSeek V3.1 - Mistral Medium 3.1 | 0.12 | 0.15 | [-0.33, 0.57] | 0.80 | 0.985 | |
| (GPT-5) - (gpt-oss 120B) | -0.78 | 0.15 | [-1.23, -0.33] | -5.14 | < .001 | *** |
| (GPT-5) - Llama 4 | -1.00 | 0.15 | [-1.45, -0.55] | -6.54 | < .001 | *** |
| (GPT-5) - Mistral Medium 3.1 | -0.36 | 0.15 | [-0.81, 0.09] | -2.34 | 0.227 | |
| (gpt-oss 120B) - Llama 4 | -0.22 | 0.15 | [-0.67, 0.22] | -1.46 | 0.766 | |
| (gpt-oss 120B) - Mistral Medium 3.1 | 0.42 | 0.15 | [-0.03, 0.87] | 2.78 | 0.080 | |
| Llama 4 - Mistral Medium 3.1 | 0.64 | 0.15 | [0.19, 1.09] | 4.22 | < .001 | *** |

Table C3 Pairwise Model Comparisons (Tukey-adjusted)

### C3 Language-Level Effects

| factor | emmean | SE | CI |
|---|---|---|---|
| Afrikaans | -1.85 | 0.34 | [-2.52, -1.17] |
| Arabic | -1.63 | 0.34 | [-2.29, -0.97] |
| Brazilian Portuguese | -2.76 | 0.35 | [-3.44, -2.09] |
| Cantonese | -2.06 | 0.33 | [-2.72, -1.41] |
| Czech | -1.73 | 0.33 | [-2.38, -1.08] |
| Dutch | -1.55 | 0.34 | [-2.21, -0.89] |
| Hebrew | -1.68 | 0.33 | [-2.34, -1.03] |
| Hindi | -1.24 | 0.33 | [-1.89, -0.60] |
| Japanese | -2.14 | 0.34 | [-2.80, -1.48] |
| Korean | -1.71 | 0.34 | [-2.38, -1.03] |
| Mandarin | -3.37 | 0.36 | [-4.07, -2.67] |
| Russian | -2.37 | 0.35 | [-3.06, -1.68] |
| Spanish | -1.68 | 0.33 | [-2.33, -1.03] |
| Swahili | -1.67 | 0.33 | [-2.32, -1.01] |
| Urdu | -1.52 | 0.35 | [-2.20, -0.84] |

Table C4 Estimated Marginal Means by Language

| contrast | estimate | SE | CI | z_ratio | p_value | Significance |
|---|---|---|---|---|---|---|
| Afrikaans - Arabic | -0.22 | 0.48 | [-1.84, 1.41] | -0.45 | 1.000 | |
| Afrikaans - Brazilian Portuguese | 0.92 | 0.48 | [-0.73, 2.56] | 1.89 | 0.857 | |
| Afrikaans - Cantonese | 0.22 | 0.48 | [-1.40, 1.84] | 0.45 | 1.000 | |
| Afrikaans - Czech | -0.12 | 0.48 | [-1.73, 1.49] | -0.25 | 1.000 | |
| Afrikaans - Dutch | -0.29 | 0.48 | [-1.91, 1.33] | -0.62 | 1.000 | |
| Afrikaans - Hebrew | -0.16 | 0.48 | [-1.78, 1.46] | -0.34 | 1.000 | |
| Afrikaans - Hindi | -0.60 | 0.47 | [-2.21, 1.00] | -1.28 | 0.995 | |
| Afrikaans - Japanese | 0.29 | 0.48 | [-1.33, 1.91] | 0.61 | 1.000 | |
| Afrikaans - Korean | -0.14 | 0.48 | [-1.78, 1.50] | -0.29 | 1.000 | |
| Afrikaans - Mandarin | 1.53 | 0.49 | [-0.15, 3.20] | 3.09 | 0.120 | |
| Afrikaans - Russian | 0.53 | 0.49 | [-1.14, 2.19] | 1.07 | 0.999 | |
| Afrikaans - Spanish | -0.17 | 0.47 | [-1.77, 1.44] | -0.35 | 1.000 | |
| Afrikaans - Swahili | -0.18 | 0.48 | [-1.80, 1.44] | -0.37 | 1.000 | |
| Afrikaans - Urdu | -0.33 | 0.49 | [-1.98, 1.32] | -0.67 | 1.000 | |
| Arabic - Brazilian Portuguese | 1.13 | 0.48 | [-0.48, 2.75] | 2.38 | 0.532 | |
| Arabic - Cantonese | 0.43 | 0.47 | [-1.16, 2.03] | 0.92 | 1.000 | |
| Arabic - Czech | 0.10 | 0.47 | [-1.49, 1.68] | 0.20 | 1.000 | |
| Arabic - Dutch | -0.08 | 0.47 | [-1.67, 1.51] | -0.17 | 1.000 | |
| Arabic - Hebrew | 0.05 | 0.47 | [-1.53, 1.64] | 0.12 | 1.000 | |
| Arabic - Hindi | -0.39 | 0.46 | [-1.96, 1.18] | -0.84 | 1.000 | |
| Arabic - Japanese | 0.51 | 0.47 | [-1.09, 2.11] | 1.08 | 0.999 | |
| Arabic - Korean | 0.07 | 0.48 | [-1.54, 1.69] | 0.16 | 1.000 | |
| Arabic - Mandarin | 1.74 | 0.49 | [0.08, 3.41] | 3.56 | 0.029 | * |
| Arabic - Russian | 0.74 | 0.49 | [-0.91, 2.39] | 1.52 | 0.973 | |
| Arabic - Spanish | 0.05 | 0.46 | [-1.52, 1.62] | 0.11 | 1.000 | |
| Arabic - Swahili | 0.04 | 0.47 | [-1.56, 1.64] | 0.08 | 1.000 | |
| Arabic - Urdu | -0.11 | 0.48 | [-1.74, 1.52] | -0.23 | 1.000 | |
| Brazilian Portuguese - Cantonese | -0.70 | 0.48 | [-2.32, 0.92] | -1.47 | 0.980 | |
| Brazilian Portuguese - Czech | -1.04 | 0.47 | [-2.65, 0.57] | -2.19 | 0.672 | |
| Brazilian Portuguese - Dutch | -1.21 | 0.48 | [-2.83, 0.40] | -2.54 | 0.409 | |
| Brazilian Portuguese - Hebrew | -1.08 | 0.48 | [-2.70, 0.54] | -2.27 | 0.616 | |
| Brazilian Portuguese - Hindi | -1.52 | 0.47 | [-3.12, 0.08] | -3.23 | 0.082 | |
| Brazilian Portuguese - Japanese | -0.63 | 0.48 | [-2.24, 0.99] | -1.31 | 0.993 | |
| Brazilian Portuguese - Korean | -1.06 | 0.48 | [-2.70, 0.58] | -2.19 | 0.672 | |
| Brazilian Portuguese - Mandarin | 0.61 | 0.49 | [-1.07, 2.29] | 1.23 | 0.996 | |
| Brazilian Portuguese - Russian | -0.39 | 0.49 | [-2.06, 1.27] | -0.80 | 1.000 | |
| Brazilian Portuguese - Spanish | -1.08 | 0.47 | [-2.69, 0.52] | -2.29 | 0.596 | |
| Brazilian Portuguese - Swahili | -1.10 | 0.48 | [-2.72, 0.53] | -2.29 | 0.596 | |
| Brazilian Portuguese - Urdu | -1.25 | 0.49 | [-2.90, 0.41] | -2.56 | 0.398 | |
| Cantonese - Czech | -0.34 | 0.47 | [-1.92, 1.25] | -0.72 | 1.000 | |
| Cantonese - Dutch | -0.51 | 0.47 | [-2.10, 1.08] | -1.09 | 0.999 | |
| Cantonese - Hebrew | -0.38 | 0.47 | [-1.97, 1.21] | -0.81 | 1.000 | |
| Cantonese - Hindi | -0.82 | 0.46 | [-2.40, 0.76] | -1.76 | 0.912 | |
| Cantonese - Japanese | 0.08 | 0.47 | [-1.52, 1.67] | 0.16 | 1.000 | |
| Cantonese - Korean | -0.36 | 0.48 | [-1.97, 1.26] | -0.75 | 1.000 | |
| Cantonese - Mandarin | 1.31 | 0.49 | [-0.34, 2.96] | 2.69 | 0.310 | |
| Cantonese - Russian | 0.31 | 0.48 | [-1.33, 1.95] | 0.64 | 1.000 | |
| Cantonese - Spanish | -0.38 | 0.47 | [-1.96, 1.20] | -0.82 | 1.000 | |
| Cantonese - Swahili | -0.39 | 0.47 | [-1.99, 1.20] | -0.84 | 1.000 | |
| Cantonese - Urdu | -0.54 | 0.48 | [-2.17, 1.08] | -1.13 | 0.999 | |
| Czech - Dutch | -0.17 | 0.47 | [-1.75, 1.41] | -0.37 | 1.000 | |
| Czech - Hebrew | -0.04 | 0.47 | [-1.62, 1.54] | -0.09 | 1.000 | |
| Czech - Hindi | -0.48 | 0.46 | [-2.05, 1.08] | -1.05 | 0.999 | |

Table C5 Pairwise Language Comparisons (Tukey-adjusted)

| Czech - Japanese | 0.41 | 0.47 | [-1.17, 2.00] | 0.88 | 1.000 | |
|---|---|---|---|---|---|---|
| Czech - Korean | -0.02 | 0.47 | [-1.63, 1.59] | -0.04 | 1.000 | |
| Czech - Mandarin | 1.65 | 0.49 | [0.00, 3.30] | 3.39 | 0.050 | * |
| Czech - Russian | 0.65 | 0.48 | [-0.99, 2.28] | 1.34 | 0.992 | |
| Czech - Spanish | -0.05 | 0.46 | [-1.61, 1.52] | -0.10 | 1.000 | |
| Czech - Swahili | -0.06 | 0.47 | [-1.64, 1.53] | -0.12 | 1.000 | |
| Czech - Urdu | -0.21 | 0.48 | [-1.82, 1.41] | -0.43 | 1.000 | |
| Dutch - Hebrew | 0.13 | 0.47 | [-1.46, 1.72] | 0.28 | 1.000 | |
| Dutch - Hindi | -0.31 | 0.46 | [-1.88, 1.26] | -0.67 | 1.000 | |
| Dutch - Japanese | 0.59 | 0.47 | [-1.01, 2.18] | 1.25 | 0.996 | |
| Dutch - Korean | 0.15 | 0.48 | [-1.46, 1.77] | 0.32 | 1.000 | |
| Dutch - Mandarin | 1.82 | 0.49 | [0.17, 3.48] | 3.73 | 0.016 | * |
| Dutch - Russian | 0.82 | 0.48 | [-0.82, 2.46] | 1.69 | 0.936 | |
| Dutch - Spanish | 0.13 | 0.46 | [-1.45, 1.70] | 0.28 | 1.000 | |
| Dutch - Swahili | 0.12 | 0.47 | [-1.48, 1.71] | 0.25 | 1.000 | |
| Dutch - Urdu | -0.03 | 0.48 | [-1.66, 1.59] | -0.07 | 1.000 | |
| Hebrew - Hindi | -0.44 | 0.46 | [-2.01, 1.13] | -0.95 | 1.000 | |
| Hebrew - Japanese | 0.45 | 0.47 | [-1.14, 2.05] | 0.97 | 1.000 | |
| Hebrew - Korean | 0.02 | 0.48 | [-1.59, 1.63] | 0.04 | 1.000 | |
| Hebrew - Mandarin | 1.69 | 0.49 | [0.03, 3.34] | 3.46 | 0.040 | * |
| Hebrew - Russian | 0.69 | 0.48 | [-0.95, 2.33] | 1.42 | 0.986 | |
| Hebrew - Spanish | 0.00 | 0.46 | [-1.58, 1.57] | -0.01 | 1.000 | |
| Hebrew - Swahili | -0.02 | 0.47 | [-1.61, 1.58] | -0.03 | 1.000 | |
| Hebrew - Urdu | -0.17 | 0.48 | [-1.79, 1.46] | -0.35 | 1.000 | |
| Hindi - Japanese | 0.90 | 0.47 | [-0.68, 2.47] | 1.93 | 0.840 | |
| Hindi - Korean | 0.46 | 0.47 | [-1.13, 2.06] | 0.98 | 1.000 | |
| Hindi - Mandarin | 2.13 | 0.48 | [0.49, 3.77] | 4.40 | 0.001 | ** |
| Hindi - Russian | 1.13 | 0.48 | [-0.50, 2.76] | 2.35 | 0.551 | |
| Hindi - Spanish | 0.44 | 0.46 | [-1.12, 1.99] | 0.95 | 1.000 | |
| Hindi - Swahili | 0.43 | 0.47 | [-1.15, 2.01] | 0.92 | 1.000 | |
| Hindi - Urdu | 0.28 | 0.47 | [-1.33, 1.89] | 0.58 | 1.000 | |
| Japanese - Korean | -0.43 | 0.48 | [-2.05, 1.18] | -0.91 | 1.000 | |
| Japanese - Mandarin | 1.23 | 0.49 | [-0.42, 2.89] | 2.53 | 0.416 | |
| Japanese - Russian | 0.23 | 0.48 | [-1.41, 1.87] | 0.48 | 1.000 | |
| Japanese - Spanish | -0.46 | 0.47 | [-2.04, 1.12] | -0.98 | 1.000 | |
| Japanese - Swahili | -0.47 | 0.47 | [-2.07, 1.13] | -1.00 | 1.000 | |
| Japanese - Urdu | -0.62 | 0.48 | [-2.25, 1.01] | -1.29 | 0.994 | |
| Korean - Mandarin | 1.67 | 0.49 | [-0.01, 3.34] | 3.38 | 0.052 | |
| Korean - Russian | 0.67 | 0.49 | [-1.00, 2.33] | 1.36 | 0.991 | |
| Korean - Spanish | -0.02 | 0.47 | [-1.63, 1.58] | -0.05 | 1.000 | |
| Korean - Swahili | -0.04 | 0.48 | [-1.65, 1.58] | -0.08 | 1.000 | |
| Korean - Urdu | -0.19 | 0.49 | [-1.83, 1.46] | -0.38 | 1.000 | |
| Mandarin - Russian | -1.00 | 0.50 | [-2.69, 0.69] | -2.01 | 0.793 | |
| Mandarin - Spanish | -1.69 | 0.48 | [-3.34, -0.05] | -3.50 | 0.036 | * |
| Mandarin - Swahili | -1.70 | 0.49 | [-3.36, -0.05] | -3.50 | 0.035 | * |
| Mandarin - Urdu | -1.86 | 0.50 | [-3.54, -0.17] | -3.74 | 0.015 | * |
| Russian - Spanish | -0.69 | 0.48 | [-2.32, 0.94] | -1.44 | 0.984 | |
| Russian - Swahili | -0.70 | 0.48 | [-2.34, 0.94] | -1.45 | 0.982 | |
| Russian - Urdu | -0.85 | 0.49 | [-2.52, 0.82] | -1.73 | 0.923 | |
| Spanish - Swahili | -0.01 | 0.47 | [-1.59, 1.57] | -0.02 | 1.000 | |
| Spanish - Urdu | -0.16 | 0.48 | [-1.77, 1.45] | -0.34 | 1.000 | |
| Swahili - Urdu | -0.15 | 0.48 | [-1.78, 1.48] | -0.31 | 1.000 | |

Table C5 Pairwise Language Comparisons (Tukey-adjusted)

### C4 Segment Category Effects

| factor | emmean | SE | CI |
|---|---|---|---|
| cultural concepts | -2.70 | 0.10 | [-2.90, -2.50] |
| holidays | -3.02 | 0.14 | [-3.28, -2.75] |
| idioms | -1.15 | 0.14 | [-1.43, -0.88] |
| puns | -0.85 | 0.13 | [-1.10, -0.60] |

Table C6 Estimated Marginal Means by Segment Category

| contrast | estimate | SE | CI | z_ratio | p_value | Significance |
|---|---|---|---|---|---|---|
| cultural concepts - holidays | 0.31 | 0.12 | [0.01, 0.62] | 2.67 | 0.039 | * |
| cultural concepts - idioms | -1.55 | 0.13 | [-1.88, -1.22] | -12.13 | < .001 | *** |
| cultural concepts - puns | -1.85 | 0.11 | [-2.14, -1.56] | -16.42 | < .001 | *** |
| holidays - idioms | -1.86 | 0.16 | [-2.27, -1.46] | -11.90 | < .001 | *** |
| holidays - puns | -2.16 | 0.14 | [-2.54, -1.79] | -14.95 | < .001 | *** |
| idioms - puns | -0.30 | 0.15 | [-0.69, 0.09] | -2.00 | 0.188 | |

Table C7 Pairwise Category Comparisons (Tukey-adjusted)

### C5 Inter-Rater Reliability

Inter-rater reliability (IRR) was assessed to evaluate the consistency of human ratings of translation quality across participants. Because ratings were ordinal (e.g., ranging from "very good / nearly perfect" to "serious failures") and involved multiple raters, we selected complementary reliability measures to capture different aspects of agreement.

We report Krippendorff's α (ordinal), which is designed for ordered categorical data and is robust to missing values, providing a single coefficient reflecting agreement beyond chance. We additionally report Gwet's AC2 with quadratic weights, which accounts for chance agreement while being less sensitive to prevalence and marginal distributions than Cohen's κ. Quadratic weights penalise larger disagreements more heavily, reflecting the ordered structure of the rating scale. Observed and expected agreement rates derived from AC2 are also reported to aid interpretation of reliability in terms of raw concordance.

Ratings corresponding to "segment not translated" (NA) were excluded from all IRR calculations, as they reflect missing or invalid quality judgments rather than graded assessments. IRR was computed at multiple levels, including overall agreement across all languages, models, and segment categories, as well as stratified by language, model, and segment category (cultural concepts, holidays, idioms, and puns).

IRR calculations were based on item × rater matrices constructed from the cleaned data and were implemented in R using the irr and irrCAC packages.

| metric | estimate | lower 95 ci | upper 95 ci | observed agreement | expected agreement | scope | language | model | segment category |
|---|---|---|---|---|---|---|---|---|---|
| Krippendorff_ alpha | 0.448197144 | NA | NA | NA | NA | Overall | NA | NA | NA |
| Gwet_AC1_ weighted | 0.41225 | (0.31,0.514) | NA | 0.755857523 | 0.584613092 | Overall | NA | NA | NA |
| Krippendorff_ alpha | 0.498850973 | NA | NA | NA | NA | Afrikaans | Afrikaans | NA | NA |
| Gwet_AC1_ weighted | 0.14534 | (-0.098,0.389) | NA | 0.632263084 | 0.569726302 | Afrikaans | Afrikaans | NA | NA |
| Krippendorff_ alpha | 0.551735695 | NA | NA | NA | NA | Arabic | Arabic | NA | NA |
| Gwet_AC1_ weighted | 0.61952 | (0.193,1) | NA | 0.849890557 | 0.605471591 | Arabic | Arabic | NA | NA |
| Krippendorff_ alpha | 0.354424333 | NA | NA | NA | NA | Brazilian Portuguese | Brazilian Portuguese | NA | NA |
| Gwet_AC1_ weighted | 0.56154 | (0.169,0.955) | NA | 0.83130482 | 0.615255895 | Brazilian Portuguese | Brazilian Portuguese | NA | NA |
| Krippendorff_ alpha | 0.386155192 | NA | NA | NA | NA | Cantonese | Cantonese | NA | NA |
| Gwet_AC1_ weighted | 0.58182 | (0.14,1) | NA | 0.824587744 | 0.580530558 | Cantonese | Cantonese | NA | NA |
| Krippendorff_ alpha | 0.501678657 | NA | NA | NA | NA | Czech | Czech | NA | NA |
| Gwet_AC1_ weighted | 0.41474 | (-0.014,0.843) | NA | 0.731120638 | 0.540580132 | Czech | Czech | NA | NA |
| Krippendorff_ alpha | 0.57461557 | NA | NA | NA | NA | Dutch | Dutch | NA | NA |
| Gwet_AC1_ weighted | 0.55692 | (0.085,1) | NA | 0.768596935 | 0.477734454 | Dutch | Dutch | NA | NA |
| Krippendorff_ alpha | 0.525872162 | NA | NA | NA | NA | Hebrew | Hebrew | NA | NA |
| Gwet_AC1_ weighted | 0.52 | (0.037,1) | NA | 0.788096253 | 0.558531884 | Hebrew | Hebrew | NA | NA |
| Krippendorff_ alpha | 0.269765185 | NA | NA | NA | NA | Hindi | Hindi | NA | NA |
| Gwet_AC1_ weighted | 0.64013 | (0.201,1) | NA | 0.833581517 | 0.537553424 | Hindi | Hindi | NA | NA |
| Krippendorff_ alpha | 0.476073172 | NA | NA | NA | NA | Japanese | Japanese | NA | NA |
| Gwet_AC1_ weighted | 0.42556 | (0,0.851) | NA | 0.780653592 | 0.618154078 | Japanese | Japanese | NA | NA |
| Krippendorff_ alpha | 0.512877424 | NA | NA | NA | NA | Korean | Korean | NA | NA |
| Gwet_AC1_ weighted | 0.29641 | (-0.134,0.727) | NA | 0.770063675 | 0.673197163 | Korean | Korean | NA | NA |
| Krippendorff_ alpha | 0.267738681 | NA | NA | NA | NA | Mandarin | Mandarin | NA | NA |
| Gwet_AC1_ weighted | 0.60908 | (0.341,0.877) | NA | 0.806393163 | 0.504740189 | Mandarin | Mandarin | NA | NA |
| Krippendorff_ alpha | 0.372753672 | NA | NA | NA | NA | Russian | Russian | NA | NA |
| Gwet_AC1_ weighted | 0.3071 | (-0.202,0.817) | NA | 0.745234394 | 0.63231795 | Russian | Russian | NA | NA |
| Krippendorff_ alpha | 0.375234491 | NA | NA | NA | NA | Spanish | Spanish | NA | NA |
| Gwet_AC1_ weighted | 0.52538 | (0.028,1) | NA | 0.808558288 | 0.596638428 | Spanish | Spanish | NA | NA |
| Krippendorff_ alpha | 0.485056033 | NA | NA | NA | NA | Swahili | Swahili | NA | NA |
| Gwet_AC1_ weighted | 0.24906 | (-0.315,0.813) | NA | 0.714105052 | 0.619284773 | Swahili | Swahili | NA | NA |
| Krippendorff_ alpha | 0.386782145 | NA | NA | NA | NA | Urdu | Urdu | NA | NA |
| Gwet_AC1_ weighted | 0.13664 | (-0.029,0.302) | NA | 0.641737452 | 0.585034893 | Urdu | Urdu | NA | NA |

Table C8 Inter-rater reliability statistics for segment-level MT quality ratings

| Krippendorff_ alpha | 0.362971562 | NA | NA | NA | NA | Claude Sonnet 4 | NA | Claude Sonnet 4 | NA |
|---|---|---|---|---|---|---|---|---|---|
| Gwet_AC1_ weighted | 0.42987 | (0.333,0.527) | NA | 0.778941763 | 0.612268497 | Claude Sonnet 4 | NA | Claude Sonnet 4 | NA |
| Krippendorff_ alpha | 0.591731477 | NA | NA | NA | NA | Cohere Aya Expanse 8B | NA | Cohere Aya Expanse 8B | NA |
| Gwet_AC1_ weighted | 0.22705 | (0.112,0.342) | NA | 0.730289925 | 0.651062192 | Cohere Aya Expanse 8B | NA | Cohere Aya Expanse 8B | NA |
| Krippendorff_ alpha | 0.365021429 | NA | NA | NA | NA | DeepSeek V3.1 | NA | DeepSeek V3.1 | NA |
| Gwet_AC1_ weighted | 0.23255 | (0.142,0.323) | NA | 0.709368798 | 0.621304321 | DeepSeek V3.1 | NA | DeepSeek V3.1 | NA |
| Krippendorff_ alpha | 0.390678454 | NA | NA | NA | NA | GPT-5 | NA | GPT-5 | NA |
| Gwet_AC1_ weighted | 0.42612 | (0.325,0.527) | NA | 0.778789507 | 0.614534099 | GPT-5 | NA | GPT-5 | NA |
| Krippendorff_ alpha | 0.425609176 | NA | NA | NA | NA | gpt-oss 120B | NA | gpt-oss 120B | NA |
| Gwet_AC1_ weighted | 0.11716 | (0.057,0.178) | NA | 0.688983685 | 0.647708789 | gpt-oss 120B | NA | gpt-oss 120B | NA |
| Krippendorff_ alpha | 0.492022726 | NA | NA | NA | NA | Llama 4 | NA | Llama 4 | NA |
| Gwet_AC1_ weighted | 0.1872 | (0.103,0.271) | NA | 0.715267408 | 0.649690855 | Llama 4 | NA | Llama 4 | NA |
| Krippendorff_ alpha | 0.353854999 | NA | NA | NA | NA | Mistral Medium 3.1 | NA | Mistral Medium 3.1 | NA |
| Gwet_AC1_ weighted | 0.29617 | (0.209,0.384) | NA | 0.747983902 | 0.641936198 | Mistral Medium 3.1 | NA | Mistral Medium 3.1 | NA |
| Krippendorff_ alpha | 0.441472615 | NA | NA | NA | NA | cultural concepts | NA | NA | cultural concepts |
| Gwet_AC1_ weighted | 0.34828 | (0.28,0.417) | NA | 0.745008769 | 0.608740413 | cultural concepts | NA | NA | cultural concepts |
| Krippendorff_ alpha | 0.380075728 | NA | NA | NA | NA | holidays | NA | NA | holidays |
| Gwet_AC1_ weighted | 0.40557 | (0.305,0.507) | NA | 0.733721118 | 0.552041014 | holidays | NA | NA | holidays |
| Krippendorff_ alpha | 0.404880664 | NA | NA | NA | NA | idioms | NA | NA | idioms |
| Gwet_AC1_ weighted | 0.10721 | (0.048,0.166) | NA | 0.737554455 | 0.706039385 | idioms | NA | NA | idioms |
| Krippendorff_ alpha | 0.307338984 | NA | NA | NA | NA | puns | NA | NA | puns |
| Gwet_AC1_ weighted | 0.26271 | (0.16,0.366) | NA | 0.757788673 | 0.671483717 | puns | NA | NA | puns |

Table C8 Inter-rater reliability statistics for segment-level MT quality ratings

| language | model | alpha | ac2 | pairwise_ agree | strict_ agree | n_ items | n_ raters |
|---|---|---|---|---|---|---|---|
| Arabic | Cohere Aya Expanse 8B | 0.674813037 | NA | 55.1 | 26.1 | 23 | 5 |
| Japanese | Llama 4 | 0.65637168 | NA | 56.7 | 36.8 | 20 | 5 |
| Czech | Llama 4 | 0.65207732 | NA | 53.9 | 24 | 25 | 5 |
| Hebrew | gpt-oss 120B | 0.644062377 | NA | 52.4 | 20 | 25 | 5 |
| Arabic | Llama 4 | 0.634999976 | NA | 56.1 | 24 | 25 | 5 |
| Arabic | Claude Sonnet 4 | 0.634900605 | NA | 64.7 | 36 | 25 | 5 |
| Urdu | Cohere Aya Expanse 8B | 0.617955706 | NA | 52.9 | 50 | 19 | 5 |
| Hebrew | DeepSeek V3.1 | 0.614170634 | NA | 63.8 | 40 | 25 | 5 |
| Hebrew | Cohere Aya Expanse 8B | 0.611726618 | NA | 57.3 | 17.4 | 23 | 5 |
| Japanese | Cohere Aya Expanse 8B | 0.584984776 | NA | 54.5 | 21.7 | 23 | 5 |
| Dutch | Cohere Aya Expanse 8B | 0.577023671 | NA | 51.4 | 31.6 | 24 | 5 |
| Dutch | gpt-oss 120B | 0.564995102 | NA | 57.5 | 27.3 | 25 | 5 |
| Dutch | Claude Sonnet 4 | 0.555536604 | NA | 59.5 | 36.4 | 25 | 5 |
| Cantonese | Llama 4 | 0.550254155 | NA | 45.3 | 17.4 | 23 | 5 |
| Arabic | Mistral Medium 3.1 | 0.544804854 | NA | 60 | 32 | 25 | 5 |
| Czech | gpt-oss 120B | 0.528554281 | NA | 53.2 | 34.8 | 25 | 5 |
| Hebrew | Llama 4 | 0.518538232 | NA | 49 | 25 | 24 | 5 |
| Arabic | gpt-oss 120B | 0.513416055 | NA | 50.4 | 12 | 25 | 5 |
| Korean | GPT-5 | 0.512390998 | NA | 45.7 | 8 | 25 | 5 |
| Czech | Cohere Aya Expanse 8B | 0.510046027 | NA | 47.4 | 14.3 | 22 | 5 |
| Hebrew | Claude Sonnet 4 | 0.502641466 | NA | 47.5 | 28 | 25 | 5 |
| Dutch | DeepSeek V3.1 | 0.493616221 | NA | 52.7 | 23.8 | 25 | 5 |
| Dutch | Mistral Medium 3.1 | 0.493573969 | NA | 61.8 | 40.9 | 25 | 5 |
| Korean | DeepSeek V3.1 | 0.486751851 | NA | 45 | 16 | 25 | 5 |
| Arabic | GPT-5 | 0.481986498 | NA | 59.2 | 32 | 25 | 5 |
| Korean | Cohere Aya Expanse 8B | 0.474465656 | NA | 45.6 | 20 | 25 | 5 |
| Spanish | Llama 4 | 0.47124898 | NA | 45.1 | 12.5 | 25 | 5 |
| Dutch | Llama 4 | 0.47021559 | NA | 55.2 | 28.6 | 25 | 5 |
| Czech | Claude Sonnet 4 | 0.457016233 | NA | 59.7 | 37.5 | 24 | 5 |
| Afrikaans | GPT-5 | 0.456513385 | NA | 71.4 | 47.6 | 24 | 5 |
| Russian | Cohere Aya Expanse 8B | 0.448616905 | NA | 44.4 | 20 | 25 | 5 |
| Afrikaans | gpt-oss 120B | 0.441137352 | NA | 61.1 | 36.4 | 24 | 5 |
| Cantonese | Cohere Aya Expanse 8B | 0.439844702 | NA | 38.1 | 16 | 25 | 5 |
| Korean | Llama 4 | 0.432970093 | NA | 47.8 | 8.7 | 25 | 5 |
| Czech | GPT-5 | 0.428335745 | NA | 59.6 | 36 | 25 | 5 |
| Brazilian Portu- guese | gpt-oss 120B | 0.426784937 | NA | 62 | 34.8 | 25 | 5 |
| Korean | gpt-oss 120B | 0.411637737 | NA | 43.6 | 8 | 25 | 5 |
| Russian | gpt-oss 120B | 0.402388898 | NA | 41.6 | 20 | 25 | 5 |
| Afrikaans | Cohere Aya Expanse 8B | 0.402248913 | NA | 41.7 | 27.3 | 23 | 5 |
| Brazilian Portu- guese | DeepSeek V3.1 | 0.400921077 | NA | 58.1 | 36 | 25 | 5 |
| Mandarin | Llama 4 | 0.396096645 | NA | 54.1 | 28 | 25 | 5 |
| Swahili | Llama 4 | 0.392872584 | NA | 52.9 | 30.4 | 25 | 5 |
| Czech | Mistral Medium 3.1 | 0.391101109 | NA | 55.8 | 20.8 | 25 | 5 |
| Brazilian Portu- guese | Claude Sonnet 4 | 0.389526749 | NA | 64 | 44 | 25 | 5 |

Table C9 Inter-rater reliability statistics for holistic text MT quality ratings

| Spanish | GPT-5 | 0.384567319 | NA | 48.2 | 20 | 25 | 5 |
|---|---|---|---|---|---|---|---|
| Hebrew | GPT-5 | 0.382856402 | NA | 57.6 | 28 | 25 | 5 |
| Japanese | gpt-oss 120B | 0.382726192 | NA | 51.8 | 28 | 25 | 5 |
| Russian | DeepSeek V3.1 | 0.38161071 | NA | 46.8 | 17.4 | 24 | 5 |
| Hindi | Mistral Medium 3.1 | 0.380938245 | NA | 38.5 | 8 | 25 | 5 |
| Korean | Claude Sonnet 4 | 0.377620246 | NA | 46.6 | 12 | 25 | 5 |
| Cantonese | Mistral Medium 3.1 | 0.371804013 | NA | 44.8 | 16 | 25 | 5 |
| Cantonese | GPT-5 | 0.360470042 | NA | 48.6 | 20 | 25 | 5 |
| Cantonese | gpt-oss 120B | 0.348227295 | NA | 45.5 | 20.8 | 24 | 5 |
| Brazilian Portuguese | Cohere Aya Expanse 8B | 0.347385294 | NA | 54.6 | 29.2 | 24 | 5 |
| Mandarin | Cohere Aya Expanse 8B | 0.345099047 | NA | 46.1 | 16.7 | 24 | 5 |
| Russian | Llama 4 | 0.333786874 | NA | 41.6 | 12 | 25 | 5 |
| Brazilian Portuguese | Llama 4 | 0.321316883 | NA | 51 | 20 | 25 | 5 |
| Hindi | Cohere Aya Expanse 8B | 0.318208174 | NA | 46.8 | 18.2 | 23 | 5 |
| Afrikaans | Claude Sonnet 4 | 0.306930278 | NA | 63 | 34.8 | 25 | 5 |
| Arabic | DeepSeek V3.1 | 0.306619915 | NA | 44.7 | 12 | 25 | 5 |
| Hindi | GPT-5 | 0.305698654 | NA | 48.3 | 20 | 25 | 5 |
| Spanish | Claude Sonnet 4 | 0.301239732 | NA | 47.4 | 12 | 25 | 5 |
| Czech | DeepSeek V3.1 | 0.29960442 | NA | 52.3 | 24 | 25 | 5 |
| Urdu | GPT-5 | 0.288346858 | NA | 51.7 | 29.4 | 18 | 5 |
| Urdu | Claude Sonnet 4 | 0.282773726 | NA | 67.1 | 50 | 20 | 5 |
| Spanish | gpt-oss 120B | 0.281928166 | NA | 43 | 13 | 25 | 5 |
| Mandarin | gpt-oss 120B | 0.278237639 | NA | 57.2 | 24 | 25 | 5 |
| Russian | Mistral Medium 3.1 | 0.277913363 | NA | 49.2 | 20.8 | 25 | 5 |
| Japanese | Mistral Medium 3.1 | 0.270812278 | NA | 60.1 | 33.3 | 24 | 5 |
| Swahili | Mistral Medium 3.1 | 0.270118901 | NA | 43.1 | 16 | 25 | 5 |
| Japanese | Claude Sonnet 4 | 0.263590307 | NA | 58.6 | 28 | 25 | 5 |
| Swahili | Claude Sonnet 4 | 0.2617325 | NA | 52.4 | 28 | 25 | 5 |
| Spanish | DeepSeek V3.1 | 0.257918185 | NA | 42.4 | 8 | 25 | 5 |
| Russian | Claude Sonnet 4 | 0.257401126 | NA | 55.1 | 28 | 25 | 5 |
| Spanish | Cohere Aya Expanse 8B | 0.257126966 | NA | 41.9 | 13 | 25 | 5 |
| Afrikaans | Mistral Medium 3.1 | 0.256137166 | NA | 47.8 | 25 | 24 | 5 |
| Cantonese | Claude Sonnet 4 | 0.251197556 | NA | 51.4 | 24 | 25 | 5 |
| Hindi | Llama 4 | 0.24301364 | NA | 43.5 | 16 | 25 | 5 |
| Korean | Mistral Medium 3.1 | 0.242460239 | NA | 48.1 | 12.5 | 24 | 5 |
| Hebrew | Mistral Medium 3.1 | 0.239619599 | NA | 44.8 | 20 | 25 | 5 |
| Afrikaans | DeepSeek V3.1 | 0.232375967 | NA | 55.7 | 37.5 | 25 | 5 |
| Hindi | DeepSeek V3.1 | 0.229249261 | NA | 41 | 12 | 25 | 5 |
| Brazilian Portuguese | Mistral Medium 3.1 | 0.212810949 | NA | 59 | 32 | 25 | 5 |
| Urdu | Llama 4 | 0.209767274 | NA | 54.4 | 35 | 20 | 5 |
| Cantonese | DeepSeek V3.1 | 0.208166722 | NA | 50.9 | 16 | 25 | 5 |
| Swahili | gpt-oss 120B | 0.201556295 | NA | 38.8 | 21.7 | 25 | 5 |
| Dutch | GPT-5 | 0.201535135 | NA | 52.2 | 26.1 | 25 | 5 |
| Hindi | gpt-oss 120B | 0.195261741 | NA | 41.8 | 12 | 25 | 5 |
| Spanish | Mistral Medium 3.1 | 0.194733517 | NA | 47.6 | 20 | 25 | 5 |

Table C9 Inter-rater reliability statistics for holistic text MT quality ratings

| Brazilian Portuguese | GPT-5 | 0.189900439 | NA | 64.3 | 32 | 25 | 5 |
|---|---|---|---|---|---|---|---|
| Japanese | GPT-5 | 0.181908943 | NA | 59.3 | 29.2 | 24 | 5 |
| Urdu | Mistral Medium 3.1 | 0.179128246 | NA | 54.4 | 25 | 20 | 5 |
| Japanese | DeepSeek V3.1 | 0.17029338 | NA | 47.9 | 20 | 25 | 5 |
| Swahili | GPT-5 | 0.168215451 | NA | 60.1 | 37.5 | 24 | 5 |
| Hindi | Claude Sonnet 4 | 0.165242137 | NA | 38.1 | 8 | 25 | 5 |
| Swahili | DeepSeek V3.1 | 0.139262956 | NA | 49.6 | 16 | 25 | 5 |
| Swahili | Cohere Aya Expanse 8B | 0.136498089 | NA | 73 | 33.3 | 25 | 5 |
| Urdu | gpt-oss 120B | 0.128582875 | NA | 38 | 10.5 | 20 | 5 |
| Afrikaans | Llama 4 | 0.121541552 | NA | 55.2 | 21.7 | 23 | 5 |
| Mandarin | DeepSeek V3.1 | 0.114767658 | NA | 65.1 | 35 | 20 | 5 |
| Russian | GPT-5 | 0.087226249 | NA | 48 | 24 | 25 | 5 |
| Mandarin | Mistral Medium 3.1 | 0.080766145 | NA | 60.3 | 32 | 25 | 5 |
| Mandarin | Claude Sonnet 4 | 0.072938315 | NA | 52.4 | 20 | 25 | 5 |
| Urdu | DeepSeek V3.1 | 0.017972445 | NA | 52.1 | 30 | 20 | 5 |

Table C9 Inter-rater reliability statistics for holistic text MT quality ratings